# A deep learning model to reduce agent dose for contrast-enhanced MRI of the cerebellopontine angle cistern

Yunjie Chen[1*], Rianne A. Weber[2*], Olaf M. Neve[1], Stephan R. Romeijn[1], Erik F. Hensen[3], Jelmer M. Wolterink[4], Qian Tao[5], Marius Staring[1], Berit M. Verbist[1]

1. Department of Radiology, Leiden University Medical Center, Albinusdreef 2, 2333ZG Leiden, the Netherlands
2. Department of Medical Imaging, Radboud University Medical Center, Geert Grooteplein Zuid 10, 6525 GA Nijmegen, the Netherlands
3. Department of Otorhinolaryngology and Head & Neck Surgery, Leiden University Medical Center, Albinusdreef 2, 2333ZG Leiden, the Netherlands
4. Department of Applied Mathematics, Technical Medical Center, University of Twente, Drienerlolaan 5, 7522 NB Enschede, the Netherlands
5. Department of Imaging Physics, Delft University of Technology, Mekelweg 5, 2628 CD Delft, the Netherlands

* Yunjie Chen and Rianne A. Weber contributed equally to this work

**ABSTRACT**

**Objectives** To evaluate a deep learning (DL) model for reducing the agent dose of contrast-enhanced T1-weighted MRI (T1ce) of the cerebellopontine angle (CPA) cistern.

**Materials and methods** In this multi-center retrospective study, T1 and T1ce of vestibular schwannoma (VS) patients were used to simulate low-dose T1ce with varying reductions of contrast agent dose. DL models were trained to restore standard-dose T1ce from the low-dose simulation. The image quality and segmentation performance of the DL-restored T1ce were evaluated. A head and neck radiologist was asked to rate DL-restored images in multiple aspects, including image quality and diagnostic characterization.

**Results** 203 MRI studies from 72 VS patients (mean age, 58.51 ± 14.73, 39 men) were evaluated. As the input dose increased, the structural similarity index measure of the restored T1ce increased from 0.639 ± 0.113 to 0.993 ± 0.009, and the peak signal-to-noise ratio increased from 21.6 ± 3.73 dB to 41.4 ± 4.84 dB. At 10% input dose, using DL-restored T1ce for segmentation improved the Dice from 0.673 to 0.734, the 95% Hausdorff distance from 2.38 mm to 2.07 mm, and the average surface distance from 1.00 mm to 0.59 mm. Both DL-restored T1ce from 10% and 30% input doses showed excellent images (3, interquartile range (IQR) [Q3-Q1] = 3 – 3 and 3, IQR [Q3-Q1]  = 4 – 3), with the latter being considered more informative (4, IQR [Q3-Q1] = 4 – 3).

**Conclusion** The DL model improved the image quality of low-dose MRI of the CPA cistern, which makes lesion detection and diagnostic characterization possible with 10% - 30% of the standard dose.

**Key points**
**Question** Deep learning models that aid in the reduction of contrast agent dose are not extensively evaluated for MRI of the cerebellopontine angle cistern.
**Finding** Deep learning models restored the low-dose MRI of the cerebellopontine angle cistern, yielding images sufficient for vestibular schwannoma diagnosis and management.
**Clinical relevance statement** Deep learning models make it possible to reduce the use of gadolinium-based contrast agents for contrast-enhanced MRI of the cerebellopontine angle cistern.

**Abbreviations**
VS - Vestibular schwannoma           CPA - cerebellopontine angle
DL- deep learning                    FOV - field of view
SSIM - structural similarity index measure    PSNR - peak signal-to-noise ratio
95HD - 95% Hausdorff distance        ASD - average surface distance

## 1. Introduction

Vestibular schwannomas (VSs) are benign tumors arising from the vestibulocochlear nerve. They account for 80% of cerebellopontine angle (CPA) tumors [1]. Magnetic Resonance Imaging (MRI), especially contrast-enhanced T1-weighted (T1ce) MRI with a gadolinium-based contrast agent (GBCA), plays a key role in the noninvasive detection and characterization of CPA lesions [1–3]. By accumulating in tissues with rich vascularity or interstitial space, GBCAs offer enhanced visibility of tumor lesions on MR imaging [4]. While T1ce is still considered the gold standard for VS diagnosis and postoperative assessment [5–8], the necessity of a contrast agent in both the diagnostic setting as well as during longitudinal VS management is being questioned [6, 8, 9]. After initial diagnosis, the surveillance of VS is sometimes carried out with T1 without contrast and/or T2-weighted MRI [3]. Given the growing concerns about short- and long-term toxicity [10, 11], the negative environmental impact of GBCA [12], and cost-effectiveness [9, 13], there is an increased interest in reducing the use of GBCA.

The possibilities of GBCA dose reduction have been explored mainly in the field of oncology [14–17]. To acquire and evaluate low-dose images, the patients typically underwent two separate examinations with an interval of more than 24 hours [14] or a two-step agent injection within 10 minutes [15, 16]. Preliminary research suggested that MRIs with 50%-75% of the standard dose are non-inferior to the standard-dose MRI [14, 17], and the scans with an even lower dose (15% of the standard dose) may still be equally effective in lesion detection and conspicuity [15, 16]. However, the low signal-noise-ratio (SNR) of low-dose MRI may potentially add difficulties for accurate lesion delineation and require more experienced clinical experts to interpret the images [18].

Recently, deep learning (DL) models have been increasingly introduced to restore standard-dose T1ce from low-dose MRI scans [18–23]. Although promising results have been demonstrated, most validations were carried out using a small-scale dataset, and only a limited number of vestibular schwannoma patients were included [19]. In addition, the values of dose reduction in previous studies were usually determined empirically, and the impact of different dose reductions on deep learning models is not well-examined. In this study, we aimed to develop a deep learning model to restore the image quality to standard-dose MRI of the CPA for VS diagnosis and longitudinal management. We conducted a comprehensive quantitative and qualitative validation using a low-dose T1ce dataset simulated from multi-center data with various dose reductions.

## 2. Materials and methods

This retrospective study was performed at Leiden University Medical Center, a tertiary referral center for skull base pathology in the Netherlands. The institutional review board approved the study protocol (G19.115), which granted an exemption for informed consent. The overall pipeline of this study is shown in Figure 1.

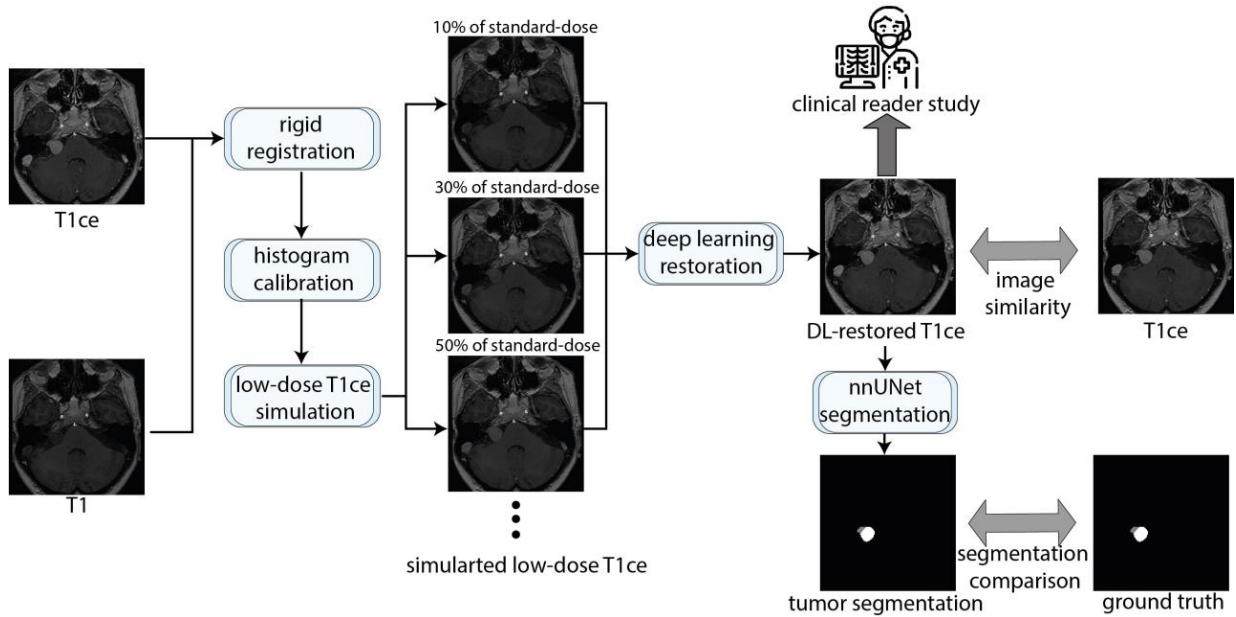

**Figure 1** The overall pipeline of the study. Nine different low-dose T1ce, ranging from 10% to 90% with 10% increments, were simulated from T1 and calibrated T1ce. The deep learning model was applied to restore T1ce from them. The DL-restored T1ce was evaluated using image similarity metrics, downstream segmentation performance, and a clinical reader study.

**Table 1** Patient characteristics. The average age of the first scan of each patient was presented as mean ± standard deviation.

| Characteristics | |
|---|---|
| Num of patients | 72 |
| Age (year) | 58.51 ± 14.73 |
| Gender (male/female) | 39/33 |
| Num of scan | 203 |
|    Intrameatal only | 89 (43.8%) |
|    Small (0-10 mm) | 35 (17.2%) |
|    Medium (11-20 mm) | 53 (26.1%) |
|    Moderately large (21–30 mm) | 17 (8.4%) |
|    Large (31-40 mm) | 9 (4.4%) |
|    Giant (>40 mm) | 0 (0.0%) |

*2.1 Patient Population and Data Preparation*

MRI studies of patients who underwent MRI of the CPA, showing a unilateral vestibular schwannoma, were collected from 33 different hospitals. Patients with other CPA pathologies, as well as multiple CPA tumors, were excluded. Every MRI study has both T1 and T1ce examinations in the transverse plane. According to the tumor classification by Kanzaki et al. [24], the proportion of intrameatal (only), small, medium, moderately large, large, and giant tumor are 43.8%, 17.2%, 26.1%, 8.4%, 4.4%, and 0.0%, respectively. The detailed characteristics and technical information of the dataset are shown in Table 1 and Table 2. The intra- and extra-meatal portions of VS were manually delineated by O.M.N., a physician with 3 years of clinical experience, who was trained in performing segmentation and, when necessary, corrected by a senior head-and-neck radiologist (B.M.V.) with 24 years of experience.

**Table 2** MRI acquisition parameters. The technical features are presented as the median value with the range in parentheses. Note that there is only one SE scan was acquired using a magnetic field of 1.0 T. (*T1ce* - contrast-enhanced T1-weighted MRI, *TE* - echo time, *TR* - repetition time, *SE* - Spin echo, *GR* - Gradient echo)

| Modality | Pulse sequence | Magnetic field (T) | TE (msec) | TR (msec) | In-plane resolution (mm) | Slice thickness (mm) |
|---|---|---|---|---|---|---|
| T1 | SE | 1.0 | 20.0 | 546.9 | 0.45 x 0.45 | 3.0 |
| | | 1.5 | 10.0 (6.0 - 24.0) | 550.0 (370.8 - 698.9) | 0.37 x 0.37 (0.32 x 0.32 − 0.86 x 0.86) | 3.0 (1.0 − 6.0) |
| | | 3.0 | 9.0 (7.0 - 16.0) | 750.0 (400.0 − 958.4) | 0.35 x 0.35 (0.27 x 0.27 − 0.66 x 0.66) | 1.0 (1.0-3.0) |
| | GR | 1.0 | 6.9 (6.7 - 6.9) | 26.0 (23.0 − 30.0) | 0.47 x 0.47 (0.35 x 0.35 − 0.90 x 0.90) | 1.4 (1.2 − 3.0) |
| | | 1.5 | 4.6 (2.4 - 5.6) | 25.0 (8.6 − 1690.0) | 0.59 x 0.59 (0.39 x 0.39 − 1.0 x 1.0) | 1.4 (1.0 − 6.0) |
| | | 3.0 | 3.7 (2.2 - 4.7) | 20.0 (8.1 − 25.0) | 0.63 x 0.63 (0.33 x 0.33 − 0.94 x 0.94) | 2.0 (0.6 − 3.0) |
| T1ce | SE | 1.0 | 20.0 | 546.9 | 0.45 x 0.45 | 3.0 |
| | | 1.5 | 11.0 (6.0 - 24.0) | 550.0 (370.8 − 698.9) | 0.37 x 0.37 (0.32 x 0.32 − 0.86 x 0.86) | 2.0 (1.0 − 5.0) |
| | | 3.0 | 9.0 (7.0 - 15.0) | 786.0 (500.1 − 958.4) | 0.35 x 0.35 (0.27 x 0.27 − 0.66 x 0.66) | 1.0 (1.0 − 3.0) |
| | GR | 1.0 | 6.9 (6.7 - 6.9) | 26.0 (23.0 − 30.0) | 0.47 x 0.47 (0.35 x 0.35 − 0.90 x 0.90) | 1.4 (1.2 − 3.0) |
| | | 1.5 | 4.6 (2.4 - 6.3) | 25.0 (8.6 − 2200.0) | 0.59 x 0.59 (0.39 x 0.39 − 1.0 x 1.0) | 1.4 (0.9 − 3.0) |
| | | 3.0 | 3.7 (2.2 − 4.7) | 20.0 (8.1 − 25.0) | 0.63 x 0.63 (0.33 x 0.33 − 0.94 − 0.94) | 2.0 (0.6- 3.0) |

Before simulating low-dose T1ce, a rigid registration was performed using Elastix [25] to ensure the spatial alignment between T1 and standard-dose T1ce. Inspired by Müller-Franzes et al. [18], we approximated the signal intensity of the low-dose images via a linear transformation. For contrast-enhanced T1, the relaxation time can be expressed as:

$$\frac{1}{T_1^{ce}} - \frac{1}{T_1} = r \cdot c,$$

where $T_1^{ce}$ and $T_1$ are the post- and pre-contrast T1 relaxation time, respectively, and $r$ and $c$ are the relaxivity and concentration of the contrast agent, respectively. When TR is short, the signal intensity can be approximated as proportional to the inverse of the T1 relaxation time [26], and thus to the concentration of the contrast agent $c$. Therefore, the low-dose images can be simulated as follows:

$$I_{low-dose} = (1 - \beta\%) \times I_{T1} + \beta\% I_{T1CE},$$

where $I_{T1}$ and $I_{T1CE}$ refer to the intensity on T1 and T1ce, respectively, and $I_{low-dose}$ refers to the intensities on contrast-enhanced T1 with $\beta\%$ of the standard dose. Intensity calibration was performed on T1ce before simulation to eliminate the influence of different intensity windows used by T1 and T1ce. Specifically, we assumed the intensity histogram of T1 and T1ce had a similar distribution, differing only in translation and scaling. We then applied translation and scaling that was determined by the second peak of each histogram (the first peak is the zero background) to the T1ce. One simulation example is shown in Figure 2 with the histograms before and after calibration. All images were normalized between -1 to 1 after simulation. Instead of using one specific proportion of dose reduction, we constructed scans of different simulated doses by varying the parameter $\beta$. As a result, nine different low-dose T1ce scans ranging from 10% to 90% of the standard dose with a 10% increment were simulated.

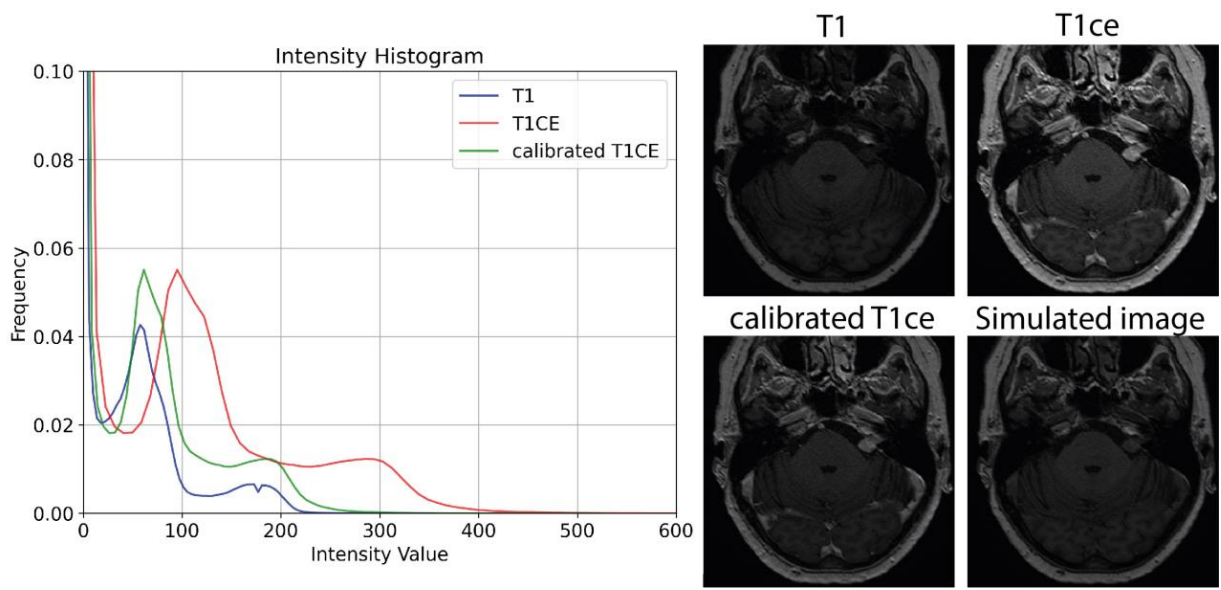

**Figure 2** An example of low-dose T1ce simulation (20% of the standard-dose). The histograms of T1, T1ce, and calibrated T1ce are shown on the left, where we can observe an intensity shift between T1ce and T1. T1, T1ce, calibrated T1CE, and a simulated low-dose image are visualized on the right.

*2.2 Deep learning model for MRI restoration*

We used the deep learning model previously developed by our group [27], which employs implicit neural representations, to restore standard-dose T1ce from simulated low-dose T1ce. Different from traditional convolutional neural network models, implicit neural representations use fully connected layers as the decoder and output 2D images as continuous fields. We trained separate models for each group of low-dose T1ce with different dose reductions, resulting in ten different models (nine models for low-dose T1ce and one model for T1 as input). All images were cropped based on the bounding box of the skull to reduce the computation and memory footprint of the model. Data augmentation, including random flipping, shifting, and cropping, was performed during training. The models were trained by a combination of $\ell_1$ reconstruction loss, adversarial loss, and $\ell_2$ regularization. We used the Adam optimizer with an initial learning rate of $1e^{-4}$, and followed the Two Time-scale Update Rule [28] to train the model. More details of the model architecture and training can be found in reference [27].

The model inference was performed with a sliding window strategy. Specifically, image patches with a size of 128 x 160 and a step size of half that (64 in height and 80 in width) were used as the input of the model. The results were aggregated using Gaussian filter smoothing. Since the enhanced regions account for only a small proportion of T1ce, we cropped the images based on the bounding box of the tumor and evaluated the image quality of these sub-images. We subsequently tested the DL-restored T1ce in a downstream segmentation task using a deep learning segmentation model previously developed by Neve et al. [29] based on nnUNet [30]. That tumor segmentation model was trained on standard-dose T1ce scans to delineate the intra- and extrameatal portions of VS. Both the training and inference were performed on a computation cluster equipped with NVIDIA Quadro RTX6000 and NVIDIA Quadro RTXA6000 GPUs using Python v3.6.8 and Pytorch v1.10.2. The full codes and experimental settings are available at *https://github.com/RianneAr/CPAMRIrestoration.*

**Table 3** Questions used in clinical reader study. The overall image quality, detectability of the lesion, and comparison of diagnostic values between images restored from different input doses were rated on a Likert scale. The enhancement pattern and diagnostic characterization were assessed in a binary manner. (*DL* - deep learning, *T1ce* - contrast-enhanced T1-weighted MRI)

| Qualitative evaluation of the DL-restored images compared to standard-dose T1ce on Likert scales | | | | | |
|---|---|---|---|---|---|
| | 1 | 2 | 3 | 4 | 5 |
| Overall image quality (vs. standard-dose T1ce) | Better | Slightly better | Similar | Slightly worse | Worse |
| Detectability of the lesion (vs. standard-dose T1ce) | Better | Slightly better | Similar | Slightly worse | Worse |
| Qualitative evaluation of the enhanced pattern of the DL-restored images | | | | | |
| Enhancement pattern | Similar | | | Different | |
| Sufficient for diagnostic characterization | Yes | | | No | |
| Qualitative comparison between images restored from 10% and 30% dose input on Likert scales | | | | | |
| | 1 | 2 | 3 | 4 | 5 |
| Diagnostic values of images restored from different input doses | 10% dose is much better | 10% dose is better | 10% dose is equivalent to 30% dose | 30% dose is better | 30% dose is much better |

*2.3 Clinical reader study*

A senior head and neck neuroradiologist with 24 years of experience rated synthetic images based on multiple criteria, as shown in Table 3. To prevent bias towards patients with more scans, we randomly select one or two scans per patient from the test set for the assessment. For each study, two DL-restored T1ce images restored from 10% and 30% input dose were displayed together with T1 and standard-dose T1ce. The reader study is not blinded, as we aim to evaluate the clinical quality of the restored images, rather than making a visual comparison between the restored images and the ground truth. The radiologist was asked to compare DL-restored MRIs with standard-dose T1ce in terms of image quality of the entire field of view (FOV) and detectability of the lesion. The results were rated on a Likert scale from 1 (DL-restored image is better than the standard-dose T1ce) to 5 (DL-restored image is worse than the standard-dose T1ce). Moreover, the radiologist was asked to identify differences in enhancement patterns, including texture, brightness, and enhancement boundary, between the real and DL-restored T1ce, and to judge if the DL-restored MRI would still be sufficient for diagnostic characterization given observed differences. Lastly, a visual comparison rating between the two DL-restored images on a Likert scale from 1 (image restored from 10% dose input is much better) to 5 (image restored from 30% input dose is much better) was performed.

*2.4 Evaluation and statistical analysis*

We evaluated the DL-restored T1ce based on the image quality within the region of interest, defined by the bounding box of VS, as well as the performance on the VS delineation task. A comparison was made between the low-dose T1ce and DL-restored T1ce. The image quality was quantified by measuring the image similarity between the region of interest and the corresponding region in standard-dose T1ce via structural similarity index measure (SSIM) and peak signal-to-noise ratio (PSNR). The performance on the VS delineation task was quantified by the Dice coefficient, the 95% Hausdorff distance (95HD), and the average surface distance (ASD). The Wilcoxon signed rank test was performed between the image quality of low-dose images and corresponding DL-restored images. *P*-values less than 0.001 were considered to indicate statistically significant differences. All analyses were performed in Python v3.6.8 with Numpy v1.21.5, SciPy v1.7.3, and scikit-learn v1.0.2.

**Table 4** Quantitative image quality of DL-restored images compared to low-dose T1ce (before restoration). The mean and standard deviation of SSIM and PSNR between the images and the standard-dose T1ce calculated on the tumor area are reported. * indicates significant difference (P < .001) between the low-dose MRI and the corresponding DL-restored MRI. (*SSIM* - structural similarity index measure, *PSNR* - peak signal-to-noise ratio, *DL* - deep learning, *T1ce* - contrast-enhanced T1-weighted MRI)

|  |  | 0 | 10 | 20 | 30 | 40 | 50 | 60 | 70 | 80 | 90 |
|---|---|---|---|---|---|---|---|---|---|---|---|
| SSIM | Before restoration | 0.562 ± 0.128 | 0.641 ± 0.107 | 0.717 ± 0.086 | 0.785 ± 0.066 | 0.845 ± 0.047 | 0.896 ± 0.032 | 0.936 ± 0.019 | 0.965 ± 0.010 | 0.985 ± 0.004 | 0.997 ± 0.001 |
|  | DL-restored | 0.639 ± 0.113* | 0.726 ± 0.104* | 0.812 ± 0.077* | 0.881 ± 0.053* | 0.921 ± 0.034* | 0.947 ± 0.024* | 0.966 ± 0.018* | 0.979 ± 0.011* | 0.987 ± 0.008 | 0.993 ± 0.009 |
| PSNR (dB) | before restoration | 19.9 ± 4.08 | 20.8 ± 4.08 | 21.8 ± 4.08 | 23.0 ± 4.08 | 24.3 ± 4.08 | 25.9 ± 4.08 | 27.9 ± 4.08 | 30.4 ± 4.08 | 33.9 ± 4.08 | 39.9 ± 4.08 |
|  | DL-restored | 21.6 ± 3.73* | 23.0 ± 3.82* | 25.6 ± 2.82* | 28.4 ± 3.10* | 30.1 ± 3.21* | 31.8 ± 3.47* | 33.3 ± 3.89* | 35.1 ± 4.29* | 37.3 ± 4.92* | 41.4 ± 4.84 |

## 3. RESULTS

A total of 203 MRI studies from 72 patients (mean age, 58.51 ± 14.73, median age, 61 (29-83), 39 men) were involved in this study. The data were divided into three groups at the patient level, resulting in a training set of 130 studies, a validation set of 25 studies, and a test set of 48 studies.

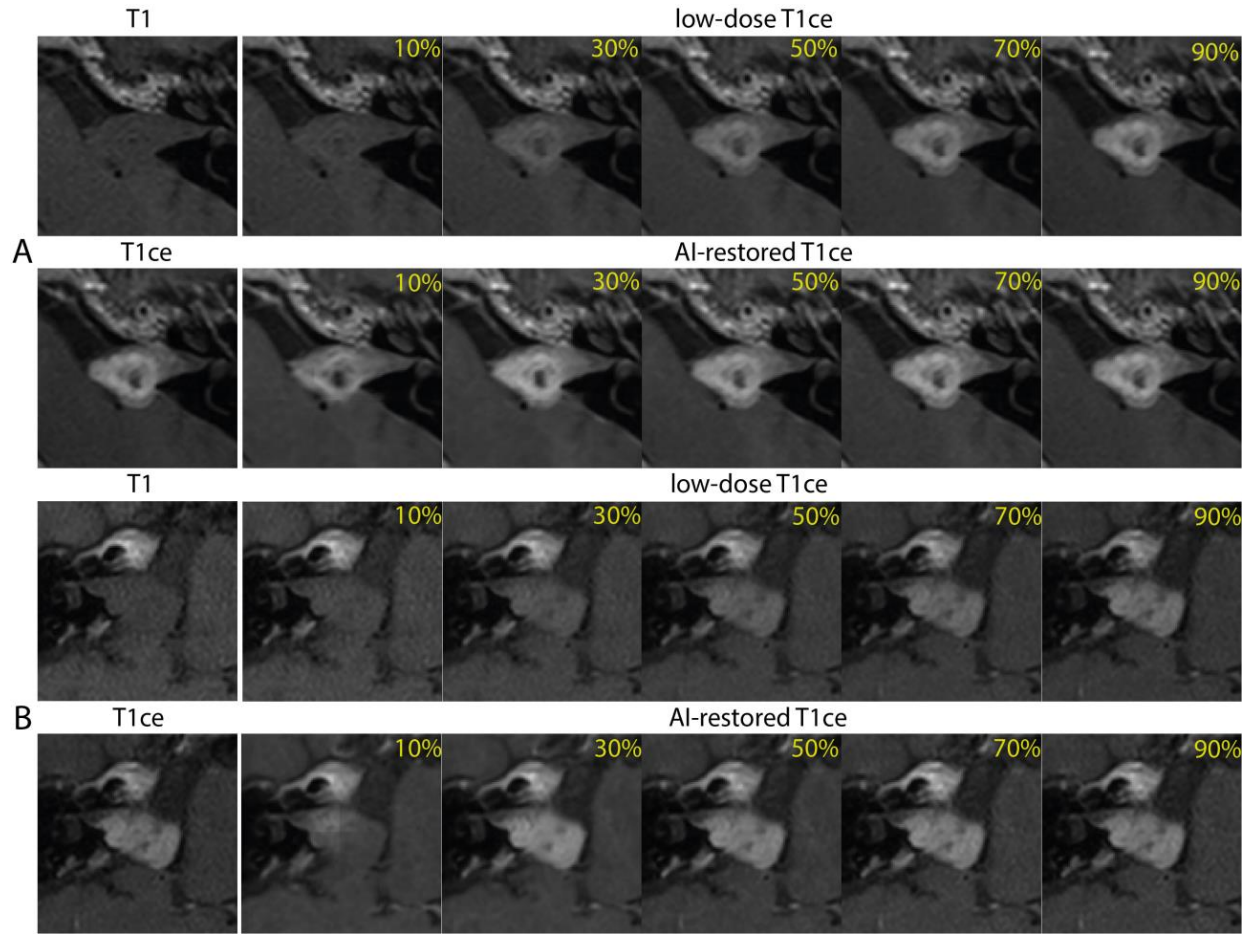

**Figure 3** Two examples (A and B) of MRI restored from different simulated low-dose T1ce. For each example, the leftmost column shows T1 (upper) and standard-dose T1ce (lower). The second to sixth columns show low-dose T1ce (upper) and their corresponding DL-restored MRI (lower). The input doses in percentage are indicated in yellow on each image. The DL-restored T1ce demonstrates a lesion enhancement, most pronounced at low input dose. However, as illustrated in B (10%), enhancement and texture restoration may be incomplete at very low input doses.

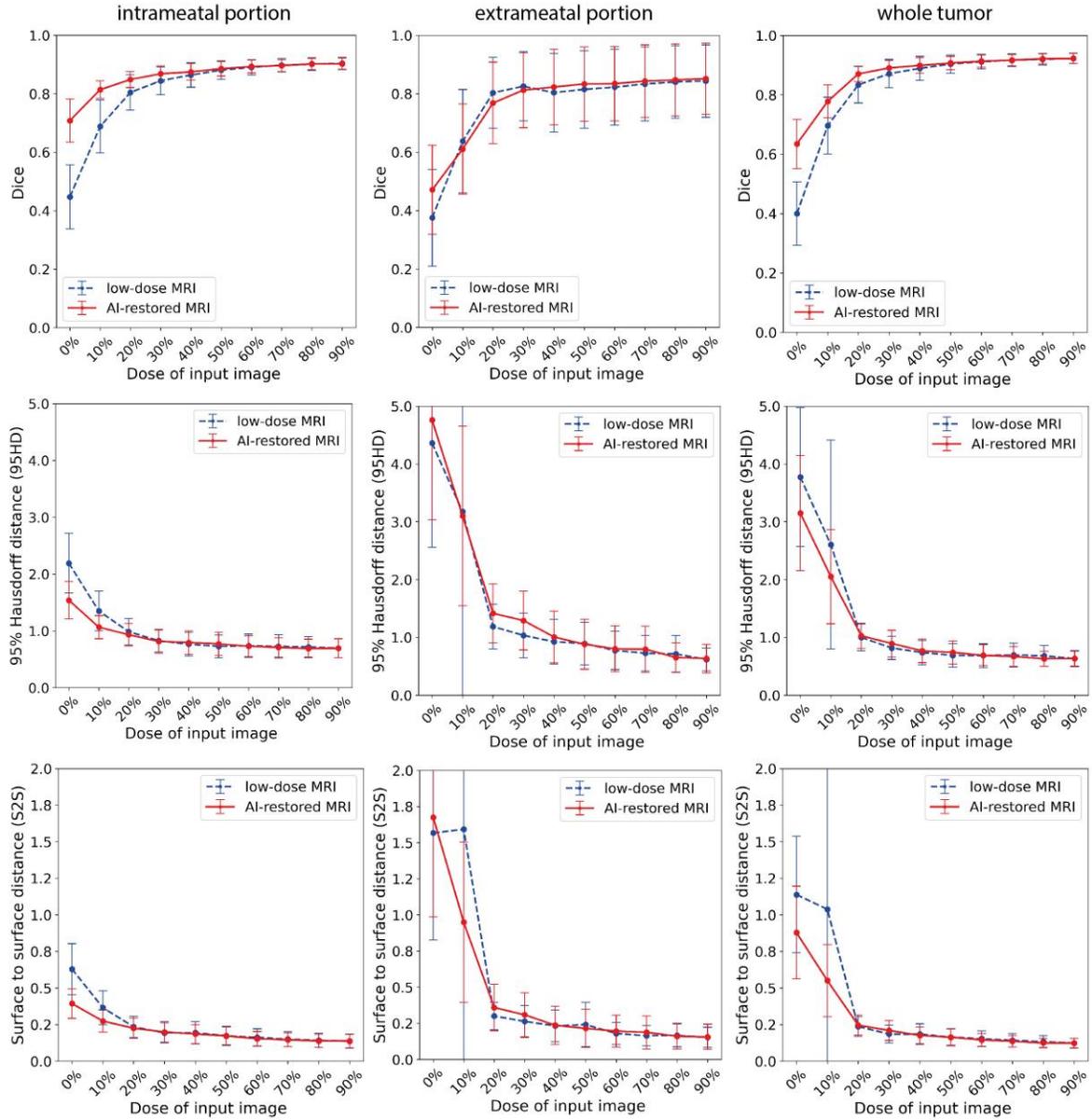

**Figure 4** Quantitative segmentation results using different dose inputs, with 95% confidence intervals shown as error bars. In each plot, the segmentation results using DL-restored MRI are indicated as a red solid line, and results using low-dose MRI are indicated as a blue dashed line. The left, middle, and right columns show plots of the intrameatal portion, extrameatal portion, and the whole tumor, respectively.

*3.1 Quantitative performance on image similarity*

The image quality of low-dose T1ce (before restoration) and DL-restored images is shown in Table 4. The DL-restoration demonstrated the most pronounced enhancement at the lowest input dose (0%) and gradually diminished as the input dose increased. Specifically, the SSIM for low-dose T1ce ranged from 0.562 ± 0.128 to 0.997 ± 0.001, and for DL-restored T1ce ranged from 0.639 ± 0.113 to 0.993 ± 0.009. The PSNR for low-dose T1ce ranged from 19.9 ± 4.08 to 39.9 ± 4.08 dB, and for DL-restored T1ce ranged from 21.6 ± 3.73 to 41.4 ± 4.84 dB. The enhancement obtained from AI restoration was statistically significant ($P < .001$) at lower dose levels (≤ 70%) but became marginal for small dose reductions (input dose ≥ 80%). Figure 3 shows examples of DL-restoration based on different input doses.

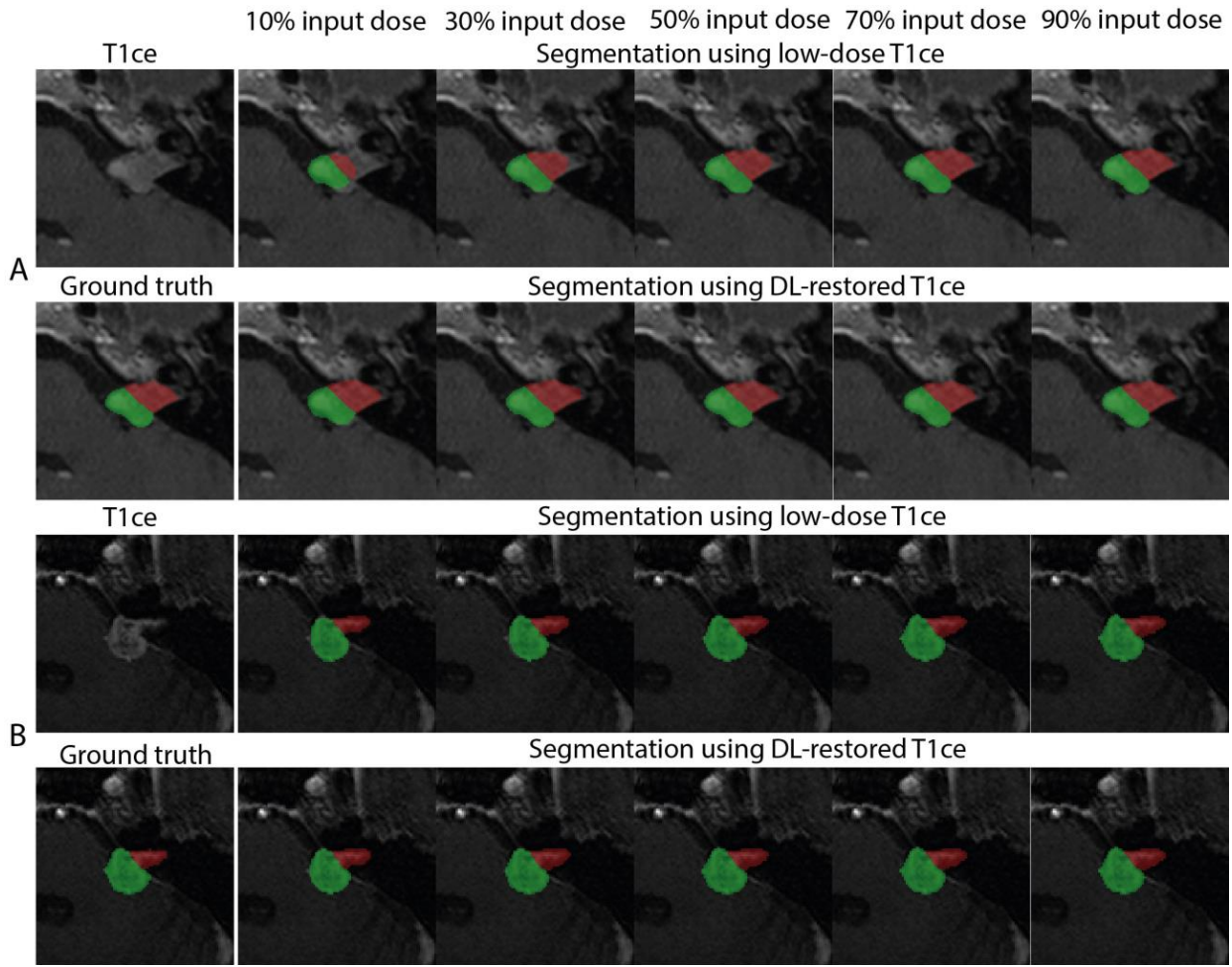

**Figure 5** Two examples of downstream segmentation experiments. The standard-dose T1ce and corresponding tumor masks are displayed in the leftmost column. Columns 2 to 6 compare segmentation results between low-dose T1ce and DL-restored T1ce. For each patient, the top row shows segmentation using low-dose MRI, and the bottom row shows segmentation using DL-restored MRI. In patient A, segmentation at low input doses underestimates the intrameatal portion (10%, 30%, and 50% input dose) and extrameatal portion (10% input dose). In contrast, the lesion was correctly segmented on restored T1ce.

## 3.2 Downstream task analysis

Figure 4 shows the quantitative results of deep learning segmentation using low-dose MRI and DL-restored MRI. Like the image quality, the enhancement from DL-restored images decreased with the increasing dose. At 10% input dose, the average Dice across the whole tumor, the intra- and extrameatal portion was 0.734 for DL-restored MRI and 0.674 for low-dose MRI. The average 95HD was 2.07 mm for DL-restored MRI and 2.38 mm for low-dose MRI. The average ASD was 0.59 mm for DL-restored MRI and 1.00 mm for low-dose MRI. At 30% input dose, the DL-restoration still improved the Dice of both the whole tumor and intrameatal portion by 0.02, but no significant differences were observed in HD95 and ASD between low-dose MRI and DL-restored MRI. From 30% upwards, the segmentation performance difference between DL-restored images and low-dose images is less pronounced. Two examples comparing segmentation using DL-restored MRI and low-dose MRI directly are shown in Figure 5.

**Table 5** The results of the clinical reader study. The median with IQR [Q3-Q1] in parentheses of the Likert scale were reported. Results of enhancement pattern and sufficiency of diagnostic characterization were presented as the number of patients with percentage in parentheses. (*T1ce* - contrast-enhanced T1-weighted MRI, *IQR* - interquartile range)

|  | Restored from 10% dose | Restored from 30% dose |
|---|---|---|
| Overall image quality (vs. standard dose T1ce) | 3 (IQR [Q3-Q1] = 3 – 3) | 3 (IQR [Q3-Q1] = 4 – 3) |
| Detectability of lesion (vs. standard dose T1ce) | 3 (IQR [Q3-Q1] = 4 – 3) | 3 (IQR [Q3-Q1] = 3 – 3) |
| Enhancement pattern (similar/different) | 7/15 (32%) | 16/8 (73%) |
| Sufficiency for diagnostic characterization (yes/no) | 10/12 (45%) | 19/3 (86%) |
| Comparison of the diagnostic values between images restored from different input doses | 4 (IQR = 4 – 3) |  |

## 3.3 Clinical reader study

Twenty-two scans were randomly selected from the test set for the clinical reader study. The results are shown in Table 5. Evaluation of the entire FOV revealed no significant differences in image quality (3, interquartile range (IQR) [Q3-Q1] = 3 – 3 vs. 3, IQR [Q3-Q1] = 4 – 3) and lesion detectability (3, IQR [Q3-Q1] = 4 – 3 vs. 3, IQR [Q3-Q1] = 3 – 3) between the DL-restored images of 10% and 30% low-dose input compared to standard-dose T1ce. However, the images restored from a 30% input dose were judged to be more informative than those from a 10% input dose, with a median rating of 4, IQR [Q3-Q1] = 4 – 3 due to better texture restoration. Specifically, 32% (7 out of 22) of the images restored from 10% input dose demonstrated similar enhancement patterns to the standard-dose T1ce, while this was the case for 73% (16 out of 22) of the images restored from 30% input dose. The differences in enhancement pattern on scans restored from 10% input dose included enhancement outside the tumor (8), (partial) lack of enhancement (6), decreased brightness (5), and cystic components being imperceptible (1). Among the six dissimilar scans that were restored from 30% input dose, four scans showed a slight overestimation of the intrameatal portion, and two scans showed different brightness in the lesion. Given that under- or overestimation of the intrameatal portion has no major impact on clinical decision-making [3], 45% of the images restored from 10% dose input were deemed sufficient for diagnosis, compared to 86% of the images restored with a 30% dose input.

## 4. DISCUSSION

In this study, we retrospectively evaluated a deep learning model to restore standard-dose contrast-enhanced MRI of the cerebellopontine angle from simulated low-dose MRI. We evaluated the restored images using multiple metrics to determine the clinical applicability of the model. Using simulated data enabled us to evaluate the proposed model with retrospective data at a large scale and explore the impact of different dose reductions on the deep learning model's performance. Our experiments showed that deep learning models can significantly improve image quality, especially when the input dose was substantially reduced. With DL-restored images, it is possible for lesion detection and further diagnostic characterization with reduced contrast doses.

In previous GBCA dose reduction studies of brain MRI, only a limited number of MRIs of the CPA were included [19, 21, 22]. For instance, of the 83 patients described by Luo et al. [19], who underwent zero-dose, 10% dose, and standard-dose brain MRI, only five had a VS. Our study, by contrast, included both the diagnostic and follow-up MRI scans of 72 vestibular schwannoma patients from 33 different medical centers, which enabled more detailed and robust enhancement evaluations related to CPA tumors. Moreover, due to the restrictions of low-dose imaging protocols, the values of dose reduction were usually determined empirically, and the scope of datasets of previous studies was generally limited. In this retrospective study, we simulated low-dose T1ce with various dose reductions and trained separate models for each group of low-dose T1ce. A comprehensive evaluation on the impact of different dose reductions on deep learning models was conducted.

Different from previous studies, we evaluated the image quality of the DL-restored T1ce based on local image similarity to the standard-dose T1ce via SSIM and PSNR. In the experiment, DL-restoration demonstrated significant improvement both qualitatively and quantitatively compared to the original low-dose T1ce. We observed that DL-restoration was most useful when the input dose was substantially reduced, and the quantitative improvements decreased with increasing input dose. This is in line with the previous study [23], in which MRI with doses ranging from 5% - 25% of the standard dose were used. As a further step, our study validated the effectiveness of the deep learning model in restoring the enhanced regions across a broader range of dose reductions.

In order to show the clinical applicability of the DL-restored images, we investigated the performance of the restored images in a downstream segmentation task. A previous study by Pasumarthi et al. [21] conducted a segmentation evaluation with a test set of limited size (N=17), and the results lack a quantitative statistical comparison. In our study, we used the state-of-the-art nnUNet model, trained using standard-dose images, for evaluation and conducted a comprehensive study. The results suggest that the AI restoration enhanced the segmentation performance, especially when the GBCA input dose was substantially reduced.

In addition to deep learning segmentation, a clinical reader study was performed to evaluate the clinical impact of DL-restored MRI. According to the radiologist, both images restored from 10% input dose and 30% input dose showed equally excellent performance in terms of overall image quality and lesion detectability. However, more different enhancement patterns were observed in images restored from a 10% input dose. This indicated that images restored from a 30% input dose demonstrated better texture restoration of the tumor, offering more information on cystic changes, which are associated with a faster growth rate, and post-therapeutic changes. It is also worth noting that there was no pronounced overestimation in the extrameatal portions of all scans, which is important for precise diameter measurements of the follow-up scans. We noticed that the radiologist sometimes rated DL-restored images higher than the real T1ce because of the high signal-to-noise ratio. This is partly due to the natural smoothing effect of the pixel-wise loss, which encourages models to produce pixel-wise averages [31], used in the DL-restoration method.

While using simulation data has the advantage of scalable dose reduction, the inherent limitation is that we did not account for the noise from dose reduction and different TR during the simulation, and did not evaluate the model on real low-dose images either. Future studies using real low-dose T1ce scans should be performed to confirm the performance of DL-restoration with the actual reduction of contrast agent. Moreover, the assumption of a linear relationship between contrast dose and enhancement does not account for other factors that may influence lesion conspicuity, such as the type of GBCA [32] and wait time after injection [33]. Although the low-dose T1ce was simulated based on standard-dose T1ce, from which we know exactly what intensity difference the contrast agent provides, follow-up studies that involve relaxivity information for model development would be promising. Additionally, in this study, the T2 scan, which is commonly also part of VS care, was not used by the DL-restoration model. We expect that the involvement of additional modalities can further improve the performance of DL-based restoration. Moreover, the proposed model has demonstrated a limited contribution to images with high input doses. We attribute this to the fact that images with higher input doses are already sufficient for tumor delineation, meaning there are no significant improvements for the model to be made. Lastly, this study was limited to a retrospective dataset. A prospective study for feasibility evaluation would be valuable before the model deployment.

In conclusion, our study has shown that a deep learning restoration model can improve the image quality of MR imaging of the CPA with reduced contrast agent doses. By using AI restoration, VS lesion detection can be done with only 10% of the standard dose, and further diagnostic characterization is possible with 30% of the standard dose. In addition, this deep learning technique could also be applied to other pathologies that are visualized with contrast-enhanced MR imaging, in order to reduce the need for contrast agents on a broader scale.

**Acknowledge**

This work was supported by the China Scholarship Council (grant 202008130140) and by an unrestricted grant of Stichting Hanarth Fonds, The Netherlands (project MLSCHWAN).

**References**

1. Lin EP, Crane BT (2017) The Management and Imaging of Vestibular Schwannomas. American Journal of Neuroradiology 38:2034–2043

2. Singh K, Singh MP, Thukral CL, et al (2015) Role of Magnetic Resonance Imaging in Evaluation of Cerebellopontine Angle Schwannomas. Indian Journal of Otolaryngology and Head and Neck Surgery 67:21–27

3. Lassaletta L, Acle Cervera L, Altuna X, et al (2024) Clinical practice guideline on the management of vestibular schwannoma. Acta Otorrinolaringologica (English Edition) 75:108–128

4. Smirniotopoulos JG, Murphy FM, Rushing EJ, et al (2007) From the archives of the AFIP: Patterns of contrast enhancement in the brain and meninges. Radiographics 27:525–551

5. Annesley-Williams DJ, Laitt RD, Jenkins JPR, et al (2001) Magnetic resonance imaging in the investigation of sensorineural hearing loss: Is contrast enhancement still necessary? Journal of Laryngology and Otology 115:14–21

6. Graffeo CS, Sivakumar W, Tavakol SA, et al (2025) Congress of Neurological Surgeons Systematic Review and Evidence-Based Guidelines Update for the Role of Imaging in the Management of Patients With Vestibular Schwannomas. Neurosurgery 00:1–5

7. Goldbrunner R, Weller M, Regis J, et al (2020) Eano guideline on the diagnosis and treatment of vestibular schwannoma. Neuro Oncol 22:31–45

8. Silva VAR, Lavinsky J, Pauna HF, et al (2023) Brazilian Society of Otology task force – Vestibular Schwannoma – evaluation and treatment. Braz J Otorhinolaryngol 89

9. Coelho DH, Tang Y, Suddarth B, Mamdani M (2018) MRI surveillance of vestibular schwannomas without contrast enhancement: Clinical and economic evaluation. Laryngoscope 128:202–209

10. Coimbra S, Rocha S, Sousa NR, et al (2024) Toxicity Mechanisms of Gadolinium and Gadolinium-Based Contrast Agents—A Review. Int J Mol Sci 25

11. Choi JW, Moon W-J (2019) Gadolinium Deposition in the Brain: Current Updates. Korean J Radiol 20:134

12. Dekker HM, Stroomberg GJ, Van der Molen AJ, Prokop M (2024) Review of strategies to reduce the contamination of the water environment by gadolinium-based contrast agents. Insights Imaging 15

13. Crowson MG, Rocke DJ, Hoang JK, et al (2017) Cost-effectiveness analysis of a non-contrast screening MRI protocol for vestibular schwannoma in patients with asymmetric sensorineural hearing loss. Neuroradiology 59:727–736

14. Liu BP, Rosenberg M, Saverio P, et al (2021) Clinical Efficacy of Reduced-Dose Gadobutrol Versus Standard-Dose Gadoterate for Contrast-Enhanced MRI of the CNS: An International Multicenter Prospective Crossover Trial (LEADER-75). American Journal of Roentgenology 217:1195–1205

15. He D, Chatterjee A, Fan X, et al (2018) Feasibility of Dynamic Contrast-Enhanced Magnetic Resonance Imaging Using Low-Dose Gadolinium: Comparative Performance with Standard Dose in Prostate Cancer Diagnosis. Invest Radiol 53:609–615

16. Pineda F, Sheth D, Abe H, et al (2019) Low-dose imaging technique (LITE) MRI: Initial experience in breast imaging. British Journal of Radiology 92:1–6

17. Melsaether AN, Kim E, Mema E, et al (2019) Preliminary study: Breast cancers can be well seen on 3T breast MRI with a half-dose of gadobutrol. Clin Imaging 58:84–89

18. Müller-Franzes G, Huck L, Arasteh ST, et al (2023) Using Machine Learning to Reduce the Need for Contrast Agents in Breast MRI through Synthetic Images. Radiology 307

19. Luo H, Zhang T, Gong NJ, et al (2021) Deep learning–based methods may minimize GBCA dosage in brain MRI. Eur Radiol 31:6419–6428

20. Gong E, Pauly JM, Wintermark M, Zaharchuk G (2018) Deep learning enables reduced gadolinium dose for contrast-enhanced brain MRI. Journal of Magnetic Resonance Imaging 48:330–340

21. Pasumarthi S, Tamir JI, Christensen S, et al (2021) A generic deep learning model for reduced gadolinium dose in contrast-enhanced brain MRI. Magn Reson Med 86:1687–1700

22. Ammari S, Bône A, Balleyguier C, et al (2022) Can Deep Learning Replace Gadolinium in Neuro-Oncology?: A Reader Study. Invest Radiol 57:99–107

23. Müller-Franzes G, Huck L, Bode M, et al (2024) Diffusion probabilistic versus generative adversarial models to reduce contrast agent dose in breast MRI. Eur Radiol Exp 8

24. Kanzaki J, Tos M, Sanna M, Moffat DA (2003) New and modified reporting systems from the consensus meeting on systems for reporting results in vestibular schwannoma. Otology and Neurotology 24:642–648

25. Klein S, Staring M, Murphy K, et al (2010) Elastix: A toolbox for intensity-based medical image registration. IEEE Trans Med Imaging 29:196–205

26. Hendrick RE (2008) Breast MRI: Fundamentals and technical aspects

27. Chen Y, Staring M, Neve OM, et al (2024) CoNeS: Conditional neural fields with shift modulation for multi-sequence MRI translation. Machine Learning for Biomedical Imaging 2:657–685

28. Heusel M, Ramsauer H, Unterthiner T, et al (2017) GANs trained by a two time-scale update rule converge to a local Nash equilibrium. Adv Neural Inf Process Syst 2017-Decem:6627–6638

29. Neve OM, Chen Y, Tao Q, et al (2022) Fully Automated 3D Vestibular Schwannoma Segmentation with and without Gadolinium-based Contrast Material: A Multicenter, Multivendor Study. Radiol Artif Intell 4

30. Isensee F, Jaeger PF, Kohl SAA, et al (2021) nnU-Net: a self-configuring method for deep learning-based biomedical image segmentation. Nat Methods 18:203–211

31. Ledig C, Theis L, Huszár F, et al (2017) Photo-Realistic Single Image Super-Resolution Using a Generative Adversarial Network. Cvpr 2:4

32. Hao J, Pitrou C, Bourrinet P (2024) A Comprehensive Overview of the Efficacy and Safety of Gadopiclenol: A New Contrast Agent for MRI of the CNS and Body. Invest Radiol 59:124–130

33. Robert P, Vives V, Grindel AL, et al (2020) Contrast-to-dose relationship of gadopiclenol, an MRI macrocyclic gadolinium-based contrast agent, compared with gadoterate, gadobenate, and gadobutrol in a rat brain tumor model. Radiology 294:117–126